\documentclass[journal]{IEEEtran}

\usepackage{mdwmath}
\usepackage{mdwtab}
\usepackage{graphicx}
\usepackage{xcolor}
\usepackage{siunitx} % For writing μm

\usepackage{cite}

\begin{document}

\title{RatLesNetv2: A Fully Convolutional Network for Rodent Brain Lesion Segmentation}

\author{Juan~Miguel~Valverde,
        Artem~Shatillo,
        Riccardo~De~Feo,
        Olli~Gr\"ohn,
        Alejandra~Sierra,
        and~Jussi~Tohka
 
\thanks{The work of J.M. Valverde was funded from the European Union's Horizon 2020 Framework Programme (Marie Sk{\l}odowska Curie grant agreement \#740264 (GENOMMED)) and R. De Feo's work was funded from Marie Sk{\l}odowska Curie grant agreement \#691110 (MICROBRADAM). We also acknowledge the Academy of Finland grants (\#275453 to A. Sierra and \#316258 to J. Tohka). Part of the computational analysis was run on the servers provided by Bioinformatics Center, University of Eastern Finland, Finland.}
\thanks{J.M. Valverde, O. Gr\"ohn, A. Sierra and J. Tohka are with AI Virtanen Institute for Molecular Sciences, University of Eastern Finland, Kuopio 70150, Finland (e-mail: juanmiguel.valverde@uef.fi, olli.grohn@uef.fi, alejandra.sierralopez@uef.fi,  jussi.tohka@uef.fi)}% <-this % stops a space
\thanks{A. Shatillo is with Charles River Discovery Services, Kuopio 70210, Finland (e-mail: artem.shatillo@crl.com)}
\thanks{R. De Feo is with Centro Fermi - Museo Storico della Fisica e Centro Studi e Ricerche Enrico Fermi, Rome 00184, Italy, with Sapienza Università di Roma, 00185 Rome, Italy, and also with AI Virtanen Institute for Molecular Sciences, University of Eastern Finland, Kuopio 70150, Finland. (e-mail: riccardo.defeo@uniroma1.it)}% <-this % stops a space
\thanks{Published in Frontiers in Neuroscience. DOI: 10.3389/FNINS.2020.610239}
}

\maketitle

\begin{abstract}
We present a fully convolutional neural network (ConvNet), named RatLesNetv2, for segmenting lesions in rodent magnetic resonance (MR) brain images. RatLesNetv2 architecture resembles an autoencoder and it incorporates residual blocks that facilitate its optimization. RatLesNetv2 is trained end to end on three-dimensional images and it requires no preprocessing. We evaluated RatLesNetv2 on an exceptionally large dataset composed of 916 T2-weighted rat brain MRI scans of 671 rats at nine different lesion stages that were used to study focal cerebral ischemia for drug development. In addition, we compared its performance with three other ConvNets specifically designed for medical image segmentation. RatLesNetv2 obtained similar to higher Dice coefficient values than the other ConvNets and it produced much more realistic and compact segmentations with notably fewer holes and lower Hausdorff distance. The Dice scores of RatLesNetv2 segmentations also exceeded inter-rater agreement of manual segmentations. In conclusion, RatLesNetv2 could be used for automated lesion segmentation, reducing human workload and improving reproducibility. RatLesNetv2 is publicly available at https://github.com/jmlipman/RatLesNetv2.
 
\end{abstract}


\section{Introduction} \label{sec1}
Rodents frequently serve as models for human brain diseases. They account for more than 80\% of the animals used in research in recent years \cite{dutta2016men}. In addition to basic research, rodent models are important in, for example,  drug discovery and the development of new treatments. In vivo imaging of rodents is used for monitoring disease progression and therapeutic response in longitudinal studies. In particular, magnetic resonance imaging (MRI) is essential in pre-clinical studies for conducting quantitative analyses due to its non-invasiveness and versatility. As an example, the quantification of brain lesions requires segmenting the lesions, and the lack of reliable tools to automate rodent brain lesion segmentation forces researchers to segment these images manually.

Manual segmentation can be prohibitively time-consuming as studies involving animals may acquire hundreds of three-dimensional (3D) images. Furthermore, the difficulty of defining lesion boundaries leads to moderate inter- and intra-rater agreement; previous studies have reported that Dice coefficients \cite{dice1945measures} between annotations made by two humans can be as low as 0.73 \cite{valverde2019automatic} or 0.79 \cite{mulder2017automated}. Moderate inter-rater agreement is caused by several factors that affect the segmentation quality, including partial volume effect, image contrast and annotator's knowledge and experience. Despite these liabilities, manual segmentation is the gold standard and a common practice among researchers who use animal models \cite{de2019towards,moraga2016imaging}.

\begin{figure}[!t]
\centering
\includegraphics[width=3.3in]{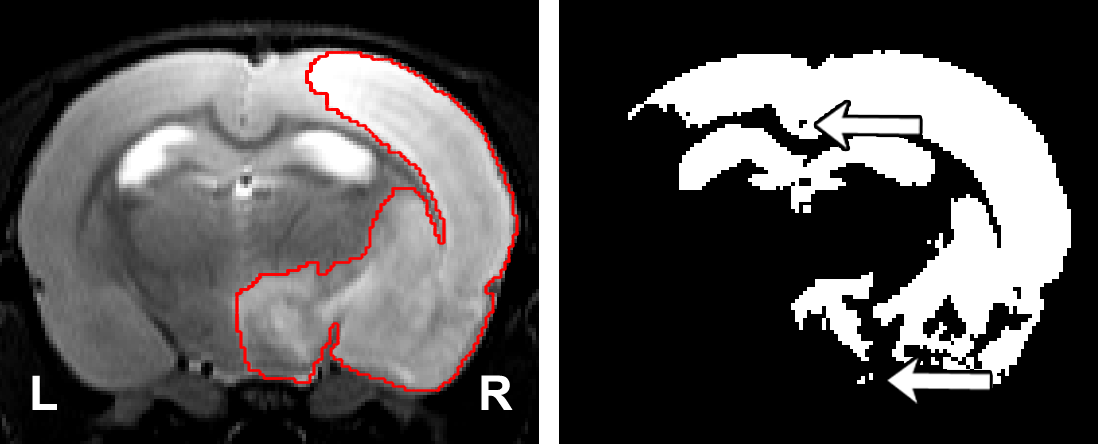}
\caption{Left: Representative lesion with its ground truth. Right: Segmentation of the lesion using thresholding where the threshold was found by maximizing the Dice coefficient with respect to the manual segmentation. The arrows indicate the presence of holes and islands (independently connected components) within and outside the mask, respectively. The hippocampus and ventricles were entirely misclassified as lesion.}
\label{fig:holes_all} % 02NOV2016/24h/13
\end{figure}

Semi-automatic methods are a faster alternative to manual segmentation. However, they fail to overcome the subjectivity of the manual segmentation, as human interaction is required. To the best of the authors' knowledge, there are only two studies that introduce and evaluate semi-automatic algorithms for rodent brain lesion segmentation. \cite{wang2007comparing} evaluated a combination of thresholding operations commonly used in the literature to segment lesions on apparent diffusion coefficient (ADC) maps and T2-weighted images. \cite{choi2018novel} first normalized the intensity values of each image with respect to the contralateral hemisphere of the brain, and they performed a series of thresholding operations to segment permanent middle cerebral artery occlusion ischemic lesions in 31 diffusion-weighted images (DWIs) of the rat brain. Both methods require the manual segmentation of the contralateral hemisphere. Additionally, these thresholding-based and other voxel-wise approaches disregard the spatial and contextual information of the images, and they are sensitive to the image modality, contrast, and possible artifacts. Pipelines that rely on thresholding operations may result in poor and inconsistent segmentation results in the form of holes within and outside the lesion mask (Fig. \ref{fig:holes_all}).

For lesion segmentation in rodent MRI, researchers have proposed a few fully-automated methods in recent years. \cite{mulder2017automated} developed a level-set-based algorithm that was tested on 121 T2-weighted mouse brain scans. However, the accuracy of their method heavily relies on the performance of other independent steps, such as registration, skull-stripping and contralateral ventricle segmentation. \cite{arnaud2018fully} derived a pipeline that detects voxels that are anomalous with respect to a reference model of healthy animals, and they evaluated the pipeline on 53 rat brain MRI maps. Nonetheless, this pipeline was specifically designed for quantitative MRI, and it expects sham-operated animals in the data set, a requirement that is not always feasible.

Deep learning, and more specifically convolutional neural networks (ConvNets), has become increasingly popular due to its competitive performance in medical image segmentation. Literature on brain lesion segmentation in MR images with ConvNets is dominated by approaches tested on human-derived data, for example,  \cite{gabr2019brain,duong2019convolutional,yang2019clci}. Despite using ConvNets, typical brain lesion segmentation approaches are multi-step, i.e., they rely on preprocessing procedures such as noise reduction, registration, skull-stripping and inhomogeneity correction. Therefore, the performance of the preprocessing steps influences the quality of the final segmentation. In contrast to human-derived data, rodent segmentation data sets are scarce and smaller in size \cite{mulder2017mri}; consequently, ConvNet-based segmentation methods benchmarked on rodent MR images are rare. An exception---not in the lesion segmentation---is \cite{roy2018deep}'s work , who developed a framework to extract brain tissue (i.e., skull-stripping) on human and mice MRI scans after traumatic brain injury.

We present RatLesNetv2, the first 3D ConvNet for segmenting rodent brain lesions in pre-clinical MR images. Our fully-automatic approach is trained end to end, requires no preprocessing, and it was validated on a large and diverse data set composed by 916 MRI rat brain scans at 9 different lesion stages from 671 rats utilized to study focal cerebral ischemia. We extend our earlier conference paper \cite{valverde2019automatic} by 1) improving our previous ConvNet \cite{valverde2019automatic} with a deeper and different architecture and providing an ablation study \cite{Meyes2019AblationSI} justifying certain architectural choices; 2) evaluating the generalization capability of our model on a considerably larger and more heterogeneous data set via Dice coefficient, compactness and Hausdorff distance under different training settings (training set size and different ground truth) and 3) making RatLesNetv2 publicly available.

We show that RatLesNetv2 generates more realistic segmentations than our previous RatLesNet, and than 3D U-Net \cite{3dunet} and VoxResNet \cite{chen2018voxresnet}, two state-of-the-art ConvNets specifically designed for medical image segmentation. Additionally, the Dice coefficients of the segmentations derived with RatLesNetv2 exceeded inter-rater agreement scores.

\section{Materials and Methods} \label{sec2}
\subsection{Data} \label{secdata}
The data set consisted of 916 MR T2-weighted brain scans of 671 adult male Wistar rats weighting between 250-300 g. The data, provided by Discovery Services site of Charles River Laboratories\footnote{https://www.criver.com/products-services/discovery-services}, were derived from 12 different studies. Transient (120 min) focal cerebral ischemia was produced by middle cerebral artery occlusion in the right hemisphere of the brain \cite{koizumimodel}. MR data acquisitions were performed at different time-points after the occlusion (for details, see Table \ref{table:dataset}). Some studies also had sham-operated animals that underwent identical surgical procedures, but without the actual occlusion. All animal experiments were conducted according to the National Institute of Health (NIH) guidelines for the care and use of laboratory animals, and approved by the National Animal Experiment Board, Finland. Multi-slice multi-echo sequence was used with the following parameters; TR = 2.5 s, 12 echo times (10-120 ms in 10 ms steps) and 4 averages in a horizontal 7T magnet. T2-weighted images were calculated as the sum of the all echoes. Eighteen coronal slices of 1 mm thickness were acquired using a field-of-view of 30x30 mm$^2$ producing 256x256 imaging matrices of resolution \SI{117x117}{\micro\meter}. No MRI preprocessing steps, such as inhomogeneity correction, artifact removal, registration or skull stripping, were applied to the T2-weighted images. Images were zero-centered and their variance was normalized to one.

\begin{table}[!t]
\renewcommand{\arraystretch}{1.3}
\caption{\normalfont{Number of scans per study segregated by lesion stage, including sham-operated animals.}}
\label{table:dataset}
\centering
\begin{tabular}{c c c c c c c c c c}
\hline \hline
Study & 2h & 24h & D3 & D7 & D14 & D21 & D28 & D35 & Shams \\
\hline
A & 12 & 12 & 0 & 0 & 0 & 0 & 0 & 0 & 24 \\
B & 0 & 46 & 0 & 0 & 0 & 0 & 0 & 0 & 3 \\
C & 0 & 59 & 0 & 0 & 0 & 0 & 0 & 0 & 1 \\
D & 0 & 162 & 0 & 0 & 0 & 0 & 0 & 0 & 4 \\
E & 0 & 0 & 0 & 0 & 0 & 0 & 0 & 20 & 1 \\
F & 0 & 33 & 30 & 0 & 30 & 0 & 27 & 0 & 46 \\
G & 0 & 0 & 0 & 53 & 0 & 0 & 0 & 0 & 12 \\
H & 0 & 45 & 0 & 0 & 0 & 0 & 0 & 0 & 0 \\
I & 0 & 0 & 64 & 0 & 0 & 0 & 62 & 0 & 0 \\
J & 0 & 32 & 0 & 0 & 0 & 0 & 0 & 0 & 0 \\
K & 0 & 17 & 0 & 0 & 0 & 0 & 0 & 0 & 0 \\
L & 0 & 0 & 41 & 0 & 0 & 40 & 0 & 0 & 40 \\
Total & 12 & 406 & 135 & 53 & 30 & 40 & 89 & 20 & 131 \\
\hline \hline
\end{tabular}
\end{table}
% A -> 02NOV2016, B -> 07MAY2015, C -> 16JUN2015, D -> 21JUL2015, E -> 03AUG2015, F -> 17NOV2015, G -> 22DEC2015, H -> 03MAY2016, I -> 08JAN2015, J -> 27JUN2017, K -> 02OCT2017, L -> 16NOV2017

The provided lesion segmentations were annotated by several trained technicians employed by Charles River. We performed an additional independent manual segmentation of the lesions on the first study that was acquired (study A, Table \ref{table:dataset}) to approximate inter-rater variability. The average Dice coefficient \cite{dice1945measures} between the two manual segmentations was 0.67 with a standard deviation of 0.12 on 2h lesions and 0.79 with a standard deviation of 0.08 on 24h lesions. The overall average was 0.73 $\pm$ 0.12. Unless stated otherwise, we used our independent segmentation as the ground truth for study A.

We produced two different train/test set divisions. 1) In the first one, the training set contained the 48 scans of the study which was used to approximate inter-rater variability (study A, Table \ref{table:dataset}) and the test set contained the remaining 868 images. The training set was further divided to training (36 images) and validation sets (12 images). This train/test division is referred to as ``\textbf{homogeneous}" and its train/validation split has the same ratio 2h/24h time-points and sham/no-sham animals. 2) The second division also contained 48 training scans and the test set contained 868 scans, but the training set was different from the homogeneous division. This division is referred to as ``\textbf{heterogeneous}" because the training set was more diverse. The training set was divided into training (40 images) and validation (8 images) set. The training and the validation sets were formed by 5 and 1 images per lesion time-point, respectively, with no images from sham-operated animals. The size of our training set was deliberately much smaller than the test set for two reasons: 1) to replicate the typical pre-clinical setting in which rodent MR images are few and 2) to create a large and representative test set.

\subsection{Convolutional neural networks}
Convolutional neural networks (ConvNets) use stacks of convolutions to transform spatially correlated data, such as images, to extract their features. The first layers of the network capture low-level information, such as edges and corners, and the final layers extract more abstract features. The number of convolutions adjusts two attributes of ConvNets: parameter number and network depth. An excessive number of parameters leads to overfitting---memorizing the training data; an insufficient number of parameters constrains the learning capability of the model. Model depth is associated with the number of times the input data is transformed, and this depth also adjusts the area that influences the prediction---the receptive field (RF). Recent approaches reduce model parameters while maintaining the RF by using more stacked convolutions of smaller kernel size \cite{szegedy2016rethinking}.

Model architectures based on U-Net \cite{ronneberger2015u} are popular in medical image segmentation tasks. In contrast to patch-based models, the input images and the generated masks are the same size, which makes U-Nets computationally more efficient to train and to evaluate. The U-Net architecture resembles an autoencoder with skip connections between the same levels of the encoder and decoder. The encoder transforms and reduces the dimensionality of the input images, and the decoder recovers the spatial information with the help of skip connections.

Skip connections also facilitate the gradient flow during back-propagation \cite{drozdzal2016importance}, but they are not sufficient to prevent the gradient of the loss to vanish, which makes the network harder to train. This is also referred as the vanishing gradient problem \cite{he2016deep}, and it particularly affects the final layers of the encoder part. Adding residual connections \cite{he2016deep} along the network alleviates the vanishing gradient problem and it also yields in faster convergence rates during the optimization \cite{drozdzal2016importance}.

\subsection{RatLesNetv2 architecture}
\begin{figure*}[!t]
\centering
\includegraphics[width=6.7in]{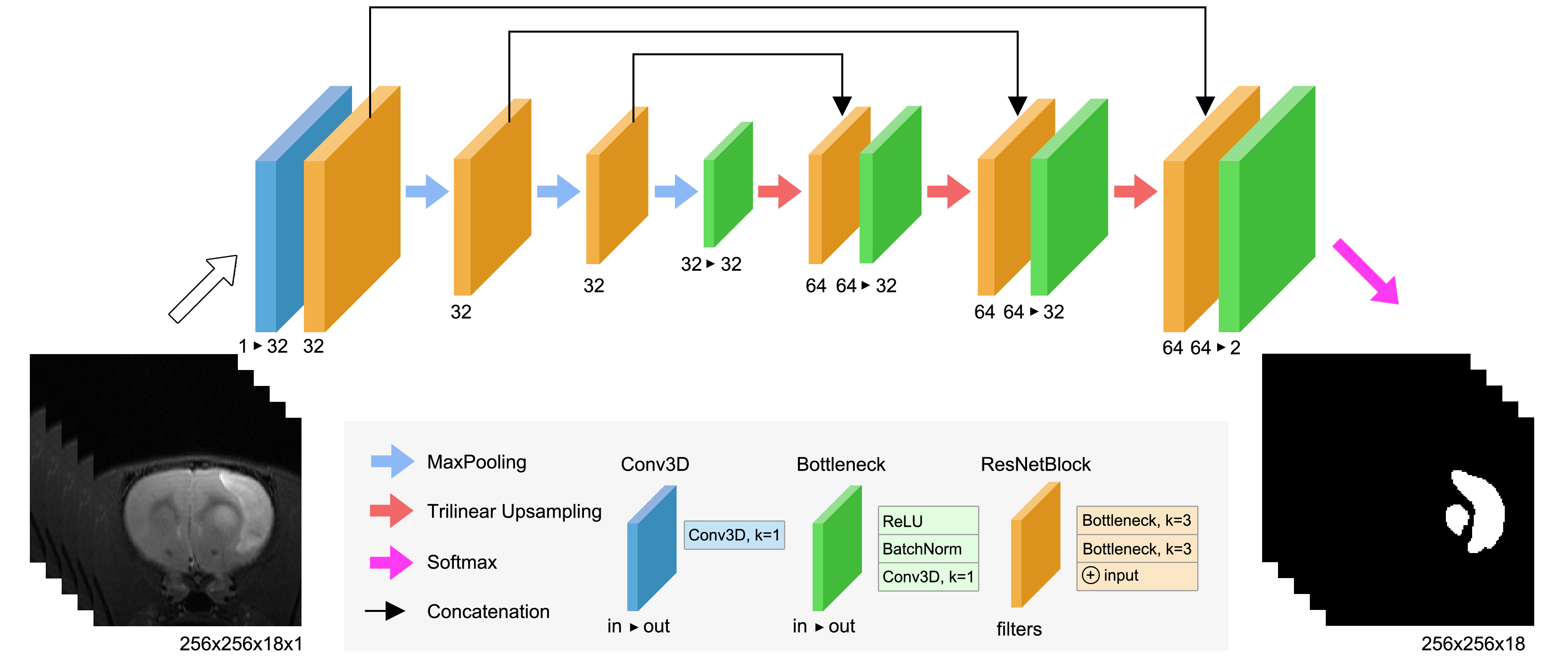}
\caption{\normalfont{RatLesNetv2 network architecture. See the text for the detailed explanation of the blocks.}}
\label{fig:architecture}
\end{figure*}

RatLesNetv2 (Fig. \ref{fig:architecture}) has three downsampling and three upsampling stages connected via skip connections. Maxpooling downsamples the data with a window size and strides of 2, and trilinear interpolation upsamples the feature maps. Bottleneck layers (Fig. \ref{fig:architecture}, green blocks) stack a ReLU activation function, a batch normalization (BatchNorm) layer \cite{ioffe2015batch} and a 3D convolution with kernel size of 1 that combines and modifies the number of channels of the feature maps from \textit{in} to \textit{out}. ResNetBlock layers (Fig. \ref{fig:architecture}, orange blocks) contain two stacks of ReLU activations, BatchNorm, and 3D convolutions with kernel size of 3. Similarly to VoxResNet \cite{chen2018voxresnet}, the input and output of each block is summed in a ResNet-style \cite{he2016deep}. The width of the blocks in the decoder is twice (64) with respect to the encoder part (32) due to the concatenation of previous layers in the same stage of the network.

At the end of the network, the probabilities $\mathbf{z} = [z_1, z_2]$  (corresponding to non-lesion and lesion labels) for each voxel are normalized by the Softmax function

\begin{equation}
    \label{eq:softmax}
    Softmax(\mathbf{z})_{i}=\frac{e^{z_{i}}}{\sum_{j=1}^{K} e^{z_{j}}},
\end{equation}
and the segmentation label is $arg max_i (q_i), i = 1,2$.

RatLesNetv2 architecture differs from our previous RatLesNet \cite{valverde2019automatic} in two aspects. First, RatLesNetv2 has one additional downsampling and upsampling level, increasing the receptive field to 76x76x76 voxels. These extra levels allows RatLesNetv2 to consider more information from a larger volume. Second, RatLesNetv2 replaces unpooling \cite{unpooling} and DenseNetBlocks \cite{densenet} with trilinear upsampling and ResNetBlocks, respectively, reducing memory usage and execution time. In contrast to VoxResNet \cite{chen2018voxresnet}, RatLesNetv2 architecture resembles an autoencoder, and RatLesNetv2 employs no transposed convolutions, reducing the number of parameters. Additionally, unlike 3D U-Net \cite{3dunet}, RatLesNetv2 uses residual blocks that reutilize previous computed feature maps and facilitate the optimization.

\subsection{Loss function}
ConvNets' parameters are optimized by minimizing a loss function that describes the difference between the predictions and the ground truth. RatLesNetv2 is optimized with Adam \cite{Kingma2014AdamAM} by minimizing cross entropy and Dice loss functions $L_{total} = L_{BCE} + L_{Dice}$. Cross entropy measures the error as the difference between distributions. Since our annotations consist of only two classes (lesion and non-lesion) we used binary cross entropy

\begin{equation}
    \label{eq:bce}
    L_{BCE}=-\frac{1}{N} \sum_{i=1}^{N} p_{i} \cdot \log(q_{i}) + (1-p_{i}) \cdot \log (1-q_{i}),
\end{equation}
where $p_{i} \in \{0, 1\}$ represents whether voxel $i$ is lesion in the ground truth and $q_{i} \in [0, 1]$ is the predicted Softmax probability of lesion class. Dice loss \cite{milletari2016v} is defined as:

\begin{equation}
    \label{eq:diceloss}
    L_{Dice}=1 - \frac{2 \sum_{i}^{N} p_{i} q_{i}}{\sum_{i}^{N} p_{i}^{2}+\sum_{i}^{N} q_{i}^{2}}.
\end{equation}
The rationale behind using Dice loss is to directly maximize the Dice coefficient, one of the metrics to assess image segmentation performance. Although the derivative of Dice loss can be unstable when its denominator is very small, the use of BatchNorm and skip connections helps during the optimization by smoothing the loss landscape \cite{li2018visualizing,santurkar2018does}.

\subsection{Post-processing}
Since our model optimizes a per-voxel loss function, small undesirable clusters of voxels may appear disconnected from the main predicted mask. These spurious clusters may be referred as ``islands" when they are separated from the largest connected component and ``holes" when they are inside the lesion mask. Figure \ref{fig:holes_all} illustrates these terms.

Small islands and holes can be removed in a final post-processing operation, yielding more realistic segmentations. Determining the maximum size of these holes and islands is, however, challenging in practice: A very small threshold will not eliminate enough small islands and a too large threshold may remove small lesions. In our experiments, we chose a threshold such that 90\% of the holes and islands in the training data were removed. More specifically, we removed holes and islands of 20 voxels or less, inside and outside the lesion masks.

\subsection{Evaluation Metrics}
We assessed the performance of each ConvNet by measuring the Dice coefficient, Hausdorff distance and compactness. In agreement with the literature \cite{Fenster2005EvaluationOS}, we argue that Dice coefficient alone is not an effective measure in rodent lesion segmentation, which is why we complemented it with the two other metrics.

\textbf{Dice coefficient}: Dice coefficient \cite{dice1945measures} is one of the most popular metrics in the field of image segmentation. It measures the overlap volume between two binary masks, typically the prediction of the model and the manually-annotated ground truth. Dice coefficient is formally described as: 

\begin{equation}
    Dice(A, B) = \frac{2|A \cap B|}{|A| + |B|},
\end{equation}
where $A$ and $B$ are the segmentation masks.

\textbf{Compactness}: Compact lesion masks are realistic and resemble human-made annotations. Compactness can be defined as the ratio between surface area ($area$) and volume of the mask ($volume$) \cite{bribiesca2008543}. More specifically, we define compactness as:

\begin{equation}
    Compactness = area^{1.5} / volume,
\end{equation}
which has a constant minimum value of $6 \sqrt\pi$ for any sphere. Compactness measure penalizes holes, islands and non-smooth borders because these increase the surface area with respect to the volume. Therefore, low compactness values that describe compact segmentations are desirable.

\textbf{Hausdorff distance}: Hausdorff distance (HD) \cite{rote1991computing} is defined as:
\begin{equation}
d(A, B)=\max \left\{\max _{a \in \partial A} \min _{b \in \partial B}|b-a|, \max _{b \in \partial B} \min _{a \in \partial A}|a-b|\right\},
\end{equation}
where $A$ and $B$ are the segmentation masks, and $\partial A$ and $\partial B$ are their respective boundary voxels. It measures the maximum distance of the ground truth surface to the closest voxel of the prediction, i.e, the largest segmentation error. Measuring Hausdorff distance in brain lesion segmentation studies is crucial since misclassifications far from the lesion boundaries are more severe. The reported Hausdorff distances were in millimeters.

Hausdorff distance and compactness values were calculated exclusively in animals with lesions. Hausdorff distance values on slightly imperfect segmentations of sham-operated animals are excessively large and distort the overall statistics. Additionally, compactness can not be calculated on empty volumes derived from scans without lesions. Voxel anisotropy was accounted for when computing HD and compactness. Finally, we assessed significance of performance difference through a paired permutation test with 10000 random iterations on the post-processed segmentations with $0.05$ as the significance threshold.

\subsection{Experimental setup}

\subsubsection{Training}
RatLesNetv2, 3D U-Net \cite{3dunet}, VoxResNet \cite{chen2018voxresnet} and RatLesNet \cite{valverde2019automatic} were optimized with Adam \cite{Kingma2014AdamAM} ($\beta_1 = 0.9, \beta_2 = 0.999, \epsilon = 10^{-8}$), starting with a learning rate of $10^{-5}$ for 700 epochs. A small set of learning rates were tested on each architecture to ensure that we used the best performing learning rate in each model. Models were randomly initialized and trained three times separately, and their performance was evaluated from the lesion masks derived with majority voting across these three independent runs. In other words, for each architecture we ensembled three independently trained models.
We confirmed that this strategy, typical to remove uncorrelated errors \cite{dietterich2000ensemble}, improves performance.

\subsubsection{Experiments}
\textbf{Performance comparison}: We optimized RatLesNetv2, 3D U-Net \cite{3dunet}, VoxResNet \cite{chen2018voxresnet} and RatLesNet \cite{valverde2019automatic} on both the homogeneous and heterogeneous data set divisions (Section \ref{secdata}) and compared their performance.

\textbf{Ablation study}: We conducted an ablation study \cite{Meyes2019AblationSI} in which we changed or removed certain parts of the model to comprehend the effects of the characteristics of RatLesNetv2 architecture. More specifically, we modified the interconnections between layers within each block, changed the number of downsampling/upsampling blocks, and increased and decreased the number of filters.

\textbf{Ground truth disparity effect}: We trained two separate RatLesNetv2 models on segmentations annotated by two different operators. This can be seen as an inter-rater variability study of the same ConvNet with disparate knowledge. We run RatLesNetv2 three times for each ground truth on the homogeneous training data, which come exclusively from the study with the two annotations (Study A, Table \ref{table:dataset}). RatLesNetv2 produced 6 sets of 868 masks $\hat{y}_{g,r}$ where $g \in \{1, 2\}$ refers to the annotator segmenting the training data and $r \in \{1, 2, 3\}$ refers to the run. First, we approximated the intra-rater variability of RatLesNetv2 by calculating the Dice coefficients among the three runs for each ground truth separately, i.e., $\{dice(\hat{y}_{g,1}, \hat{y}_{g,2}), dice(\hat{y}_{g,2}, \hat{y}_{g,3}), dice(\hat{y}_{g,1}, \hat{y}_{g,3})\}$ for $g = 1,2$. This led to two sets of 3 Dice coefficients per mask. Second, we calculated the Dice coefficient of the masks across the different ground truths $\{dice(\hat{y}_{1,i}, \hat{y}_{2,j})\}$ for $i,j = 1,2,3$ to approximate inter-rater similarity, leading to 9 Dice coefficients per mask.

\textbf{Training set size}: We optimized RatLesNetv2 with training sets of different sizes to understand the relation between training set size and generalization capability. The training sets had the same ratio of time-points, i.e., we enlarged the training sets by 1 sample per time-point. Since the lowest number of samples across time-points corresponds to 12 (2h lesions) and we want to keep at least 1 image per time-point in the test set, we produced 11 training sets $T_i$ of size $|T_i| = 8i$ for $i = 1, \ldots, 11$, where 8 is the number of lesion stages.

\subsubsection{Implementation}
RatLesNetv2 was implemented in Pytorch \cite{paszke2019pytorch} and it was run on Ubuntu 16.04 with an Intel Xeon W-2125 CPU @ 4.00GHz processor, 64 GB of memory and an NVidia GeForce GTX 1080 Ti with 11 GB of memory. RatLesNetv2 is publicly available at https://github.com/jmlipman/RatLesNetv2

\section{Results}
\subsection{Performance of RatLesNetv2}
\label{sec:performance}

Table \ref{table:comparisonvoxresnet} lists the quantitative validation results on the test set excluding sham-operated animals that typically yield Dice coefficients of 1.0. As can be seen in Table \ref{table:comparisonvoxresnet}, RatLesNetv2 produced similar or better Dice coefficients and Hausdorff distances, and more compact segmentations than the other ConvNets. The average Dice coefficients varied from 0.784 (homogeneous division) to 0.813 (heterogeneous division). Dice coefficients had a large standard deviation regardless of the architecture (from $0.15$ to $0.20$). However, note that the sample-wise difference between the Dice coefficients of RatLesNetv2 and VoxResNet had a smaller standard deviation of $0.05$, i.e., the Dice values between different networks were correlated. Table \ref{table:comparisonvoxresnet} shows that RatLesNetv2 achieved significantly better compactness values (all $p$-values $<$ 0.011) than 3D U-Net, VoxResNet and RatLesNet. Remarkably, 3D U-Net and VoxResNet produced masks with non-smooth borders and several more holes and islands, leading to less compact segmentations (see Figure \ref{fig:comparison_segemtation} and Figures in the Supplementary Material). The average compactness values of RatLesNetv2 were higher than the ground truth (20.98 $\pm$ 3.28, $p = 0.003$); this was expected as human annotators are likely to produce segmentations with excessively rounded boundaries.

\begin{figure*}[bt]
\centering
\includegraphics[width=\textwidth]{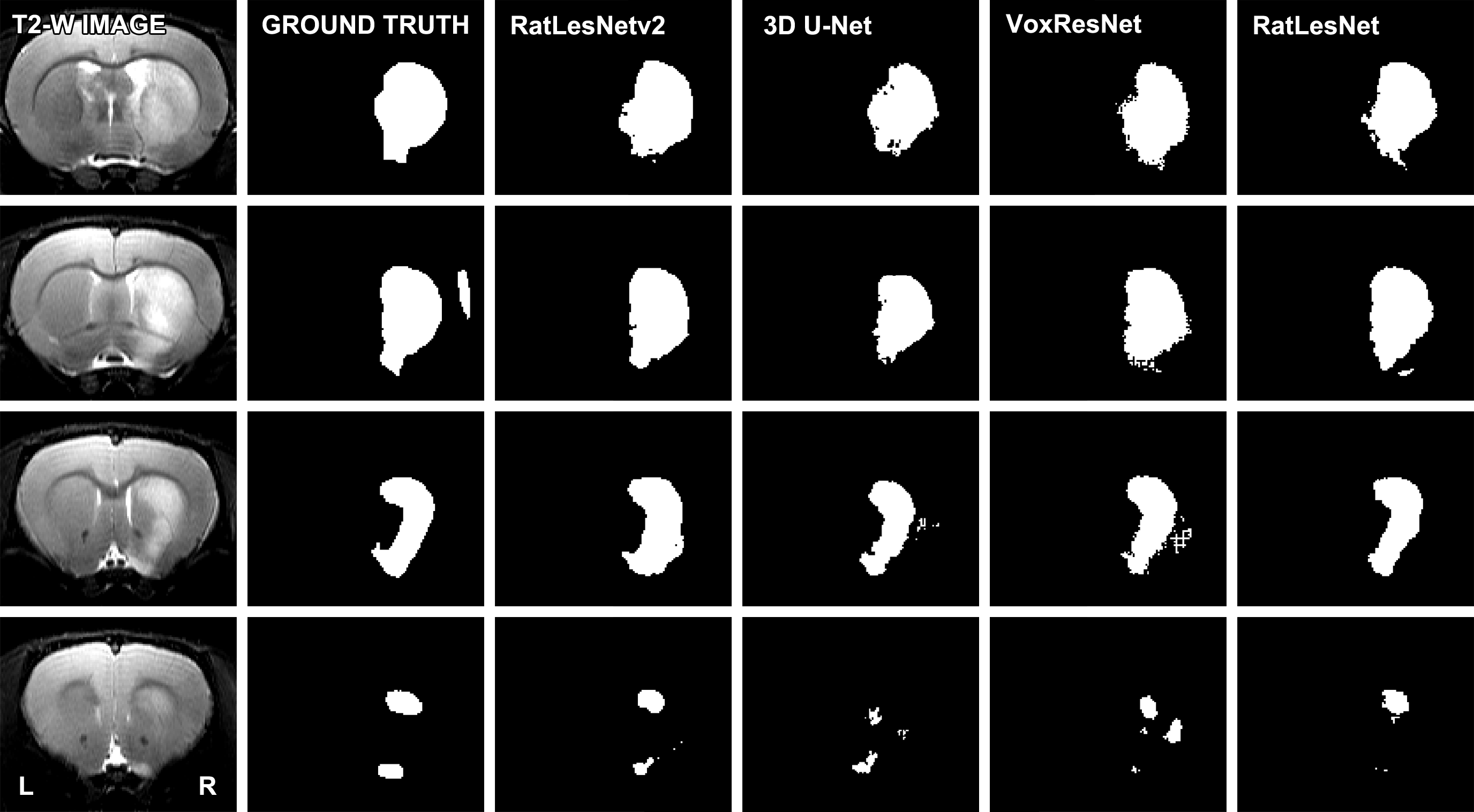}
\caption{\normalfont{Comparison of the segmentation masks of 4 consecutive slices. The depicted T2-weighted image corresponds to a typical scan, i.e., volume whose segmentation achieved the median Dice coefficient in the test set (heterogeneous division). Segmentations were not post-processed.}}
\label{fig:comparison_segemtation} % 21JUL2015_24h_450. Median Dice coeff without shams, at 2-baseline/3-allpreds
\end{figure*}

\begin{table}[!t]
\renewcommand{\arraystretch}{1.3}
\caption{\normalfont{Performance evaluation on the test set before and after post-processing.
%Masks were derived from the majority voting of 3 runs.
Top: Homogeneous division. Bottom: Heterogeneous division. Bold: Values significantly better than the other architectures (${}^{*a}P$ $=$ 0.007,${}^{*b}P$ $=$ 0.011,${}^{*c}P$ $=$ 0.005).}}
\label{table:comparisonvoxresnet}
\centering
\begin{tabular}{l c c c}
\hline \hline
Model & Dice (no shams) & Compactness & HD\\
\hline
RatLesNetv2-post & $\textbf{0.784 $\pm$ 0.18}^{*a}$ & $\textbf{29.332 $\pm$ 7.86}^{*b}$ & 3.522 $\pm$ 3.64\\
RatLesNetv2 & 0.784 $\pm$ 0.18 & 29.609 $\pm$ 8.12 & 3.687 $\pm$ 3.30\\
3D U-Net-post & 0.769 $\pm$ 0.20 & 36.741 $\pm$ 11.41 & 3.665 $\pm$ 3.81 \\
3D U-Net & 0.768 $\pm$ 0.20 & 37.599 $\pm$ 11.77 & 4.097 $\pm$ 3.69 \\
VoxResNet-post & 0.757 $\pm$ 0.19 & 37.096 $\pm$ 13.00 & 3.692 $\pm$ 3.46\\
VoxResNet & 0.757 $\pm$ 0.19 & 38.161 $\pm$ 13.62 & 4.943 $\pm$ 3.38\\
RatLesNet-post & 0.742 $\pm$ 0.18 & 35.045 $\pm$ 10.71 & 3.892 $\pm$ 2.54\\
RatLesNet & 0.741 $\pm$ 0.18 & 35.888 $\pm$ 10.76 & 4.679 $\pm$ 2.55\\
\hline
RatLesNetv2-post & 0.813 $\pm$ 0.16 & $\textbf{23.105 $\pm$ 4.58}^{*c}$ & 3.334 $\pm$ 3.34\\
RatLesNetv2  & 0.813 $\pm$ 0.16 & 23.177 $\pm$ 4.64 & 3.512 $\pm$ 3.31\\
3D U-Net-post & 0.813 $\pm$ 0.15 & 28.247 $\pm$ 5.92 & 3.099 $\pm$ 2.47\\
3D U-Net & 0.812 $\pm$ 0.15 & 28.639 $\pm$ 5.99 & 3.221 $\pm$ 2.47\\
VoxResNet-post & 0.806 $\pm$ 0.14 & 32.937 $\pm$ 10.05 & 3.585 $\pm$ 3.27\\
VoxResNet & 0.805 $\pm$ 0.14 & 33.634 $\pm$ 10.53 & 4.535 $\pm$ 3.46\\
RatLesNet-post & 0.764 $\pm$ 0.15 & 31.348 $\pm$ 9.66 & 3.218 $\pm$ 2.79\\
RatLesNet & 0.764 $\pm$ 0.15 & 31.669 $\pm$ 9.86 & 3.384 $\pm$ 2.56\\
\hline \hline
\end{tabular}
\end{table}

\begin{table}[!t]
\renewcommand{\arraystretch}{1.3}
\caption{\normalfont{Performance evaluation on the test set after post-processing segregated by lesion stage. Top: Homogeneous division. Bottom: Heterogeneous division.}}
\label{table:brokendown}
\centering
\begin{tabular}{l c c c}
\hline \hline
Time-point (scans)& Dice & Compactness & HD \\
24h (394) & 0.831 $\pm$ 0.15 & 26.539 $\pm$ 4.86 & 3.691 $\pm$ 3.53\\
D3 (135) & 0.782 $\pm$ 0.12 & 29.705 $\pm$ 8.01 & 3.067 $\pm$ 2.31\\
D7 (53) & 0.790 $\pm$ 0.11 & 40.742 $\pm$ 10.65 & 2.580 $\pm$ 2.63\\
D14 (30) & 0.735 $\pm$ 0.21 & 36.018 $\pm$ 11.07 & 3.329 $\pm$ 5.06\\
D21 (40) & 0.800 $\pm$ 0.11 & 33.797 $\pm$ 6.50 & 2.546 $\pm$ 0.92\\
D28 (89) & 0.593 $\pm$ 0.28 & 29.598 $\pm$ 7.46 & 4.238 $\pm$ 5.22\\
D35 (20) & 0.751 $\pm$ 0.23 & 31.203 $\pm$ 4.72 & 4.831 $\pm$ 5.76\\
shams (107) & 1.000 $\pm$ 0.00 & --- & --- \\
\hline
2h (6) & 0.719 $\pm$ 0.11 & 23.111 $\pm$ 2.27 & 1.920 $\pm$ 0.16\\
24h (400) & 0.826 $\pm$ 0.15 & 23.218 $\pm$ 4.67 & 3.919 $\pm$ 3.79\\
D3 (129) & 0.809 $\pm$ 0.10 & 23.376 $\pm$ 5.15 & 2.796 $\pm$ 2.24\\
D7 (47) & 0.860 $\pm$ 0.09 & 23.555 $\pm$ 3.99 & 2.439 $\pm$ 2.83\\
D14 (24) & 0.827 $\pm$ 0.19 & 21.705 $\pm$ 3.36 & 3.015 $\pm$ 5.69\\
D21 (34) & 0.877 $\pm$ 0.10 & 23.874 $\pm$ 2.38 & 2.002 $\pm$ 0.70\\
D28 (83) & 0.692 $\pm$ 0.25 & 22.147 $\pm$ 4.65 & 2.875 $\pm$ 1.93\\
D35 (14) & 0.886 $\pm$ 0.07 & 22.037 $\pm$ 2.55 & 1.700 $\pm$ 0.67\\
shams (131) & 1.000 $\pm$ 0.00 & --- & ---\\
\hline \hline
\end{tabular}
\end{table}

Post-processing had little to no effect on the average Dice coefficients, but it enhanced the final segmentation quality as it removed spurious clusters of voxels. This improvement was reflected in the reduction of compactness values and the considerable decrease of Hausdorff distances. Remarkably, the difference in the Hausdorff distances before and after post-processing was more pronounced in 3D U-Net, VoxResNet and RatLesNet.

Table \ref{table:brokendown} lists the quantitative results by lesion stage to understand the performance of RatLesNetv2 in detail. Training RatLesNetv2 on the homogeneous data division, whose training set included almost twice as many 24h lesion scans as the heterogeneous division (9 scans vs. 5 scans), led to a slight increase in the average Dice coefficient and Hausdorff distance in 24h lesion scans. However, there was no significant difference between either the Dice coefficients ($p =$ 0.057) nor Hausdorff distances ($p =$ 0.08) of the segmentations derived in the two cases. Dice coefficients, compactness values and Hausdorff distances of the segmentations produced after training on the homogeneous division deteriorated as the time-point was farther from 2h and 24h.

Training on the heterogeneous training set notably improved the average Dice coefficients and compactness values of every model (Table \ref{table:comparisonvoxresnet}) and every time-point (Table \ref{table:brokendown}) with respect to homogeneous division, except on 24h lesions. Furthermore, it decreased the standard deviation of the Dice coefficients and compactness values. RatLesNetv2 recognized animals without lesions notably well even if they were not part of the training set, providing average Dice coefficients of $1.0$ on sham-operated animals even without post-processing. Additionally, Dice coefficients on 2h lesions, 24h lesions and overall were higher than inter-rater agreement.

Ensembling three ConvNets of the same architecture optimized on the same training set led to significantly better performance scores in all cases (all $p$-values $<$ 0.007) as it discarded small segmentation inconsistencies.
This strategy increased Dice coefficients by an average of 2\% and decreased compactness and Hausdorff distances by an average of 5\% and 23\% with respect to the first run. The Dice coefficients, compactness values and Hausdorff distances from the individual images used for calculating the reported statistics are also included in the Supplementary Materials as CSV files.

\subsection{Ablation studies}

The performance scores of RatLesNetv2 after modifying its architecture during the ablation studies are reported in Table \ref{table:ablation}.

\begin{table}[!t]
\renewcommand{\arraystretch}{1.3}
\caption{\normalfont{Ablation study. Top: Homogeneous task. Bottom: Heterogeneous task. *: Same number of parameters than Baseline. Bold: baseline significantly better. Italic: baseline significantly worse (P-values $<$ 0.05).}} %Italic: 0.041, Bold: 0.047
\label{table:ablation}
\centering
\begin{tabular}{l c c c}
\hline \hline
Study & Dice (no shams) & Compactness & HD \\
\hline
Baseline & 0.784 $\pm$ 0.18 & 29.332 $\pm$ 7.86 & 3.522 $\pm$ 3.64\\
DenseNetBlock* & \textbf{0.771 $\pm$ 0.20} & \textbf{30.094 $\pm$ 8.86} & 3.692 $\pm$ 3.96\\
Halving RF & \textbf{0.754 $\pm$ 0.20} & 30.766 $\pm$ 10.57 & 3.340 $\pm$ 3.91\\
Halving RF* & \textbf{0.765 $\pm$ 0.19} & \textbf{31.867 $\pm$ 10.41} & 3.464 $\pm$ 3.53\\
Width-28 & 0.781 $\pm$ 0.18 & 30.095 $\pm$ 8.42 & 3.423 $\pm$ 2.84\\
Width-36 & \textbf{0.765 $\pm$ 0.19} & \textbf{31.620 $\pm$ 10.09} & 3.557 $\pm$ 3.73\\
\hline
Baseline & 0.813 $\pm$ 0.16 & 23.105 $\pm$ 4.58 & 3.334 $\pm$ 3.34\\
DenseNetBlock* & \textbf{0.801 $\pm$ 0.16} & \textbf{23.313 $\pm$ 5.13} & 3.093 $\pm$ 2.70\\
Halving RF & \textit{0.819 $\pm$ 0.15} & \textbf{25.226 $\pm$ 5.34} & \textbf{3.679 $\pm$ 3.12}\\
Halving RF* & \textit{0.820 $\pm$ 0.15} & \textbf{25.394 $\pm$ 5.51} & \textbf{3.719 $\pm$ 3.40}\\
Width-28 & \textbf{0.803 $\pm$ 0.17} & 22.861 $\pm$ 4.63 & \textit{2.892 $\pm$ 2.94}\\
Width-36 & \textbf{0.801 $\pm$ 0.16} & 24.036 $\pm$ 5.04 & 2.900 $\pm$ 2.84\\
\hline \hline
\end{tabular}
\end{table}

\subsubsection{DenseNetBlock}
Similarly to RatLesNet \cite{valverde2019automatic}, DenseNet-style \cite{densenet} blocks were implemented in RatLesNetv2 while keeping the same number of parameters of the baseline RatLesNetv2 model. Dice coefficients and compactness values were significantly deteriorated with respect to RatLesNetv2 baseline (all $p$-values $<$ 0.037), and Hausdorff distances increased slightly in homogeneous data division, whereas they decreased in heterogeneous division. Additionally, DenseNetBlocks demanded notably more memory due to the concatenation operation.

\subsubsection{Halving the receptive field (RF)}
The third downsampling stage of RatLesNetv2 was eliminated in order to reduce the receptive field from 72 voxels down to 36. An additional test (marked in Table \ref{table:ablation} with an *) matched the number of parameters to the baseline. The reduction of the receptive field yielded in significant improvements of the Dice coefficient and a significant deterioration of the compactness and Hausdorff distance in the heterogeneous division (all $p$-values $<$ 0.028). On the other hand, in the homogeneous division Dice coefficients and compactness values were worse than RatLesNetv2 baseline.

\subsubsection{Network Width}
We increased and decreased the number of filters of RatLesNetv2 by 4 (Table \ref{table:ablation}, Width-28 and Width-36). This modification decreased the Dice coefficients with respect to RatLesNetv2 and led to no significant difference in the Hausdorff distances. Compactness values showed contradictory results; they deteriorated in homogeneous division whereas they remained similar or slightly worse in heterogeneous division.

\subsection{On the influence of disparate ground truths}

\begin{figure*}[bt]
\centering
\includegraphics[width=\textwidth]{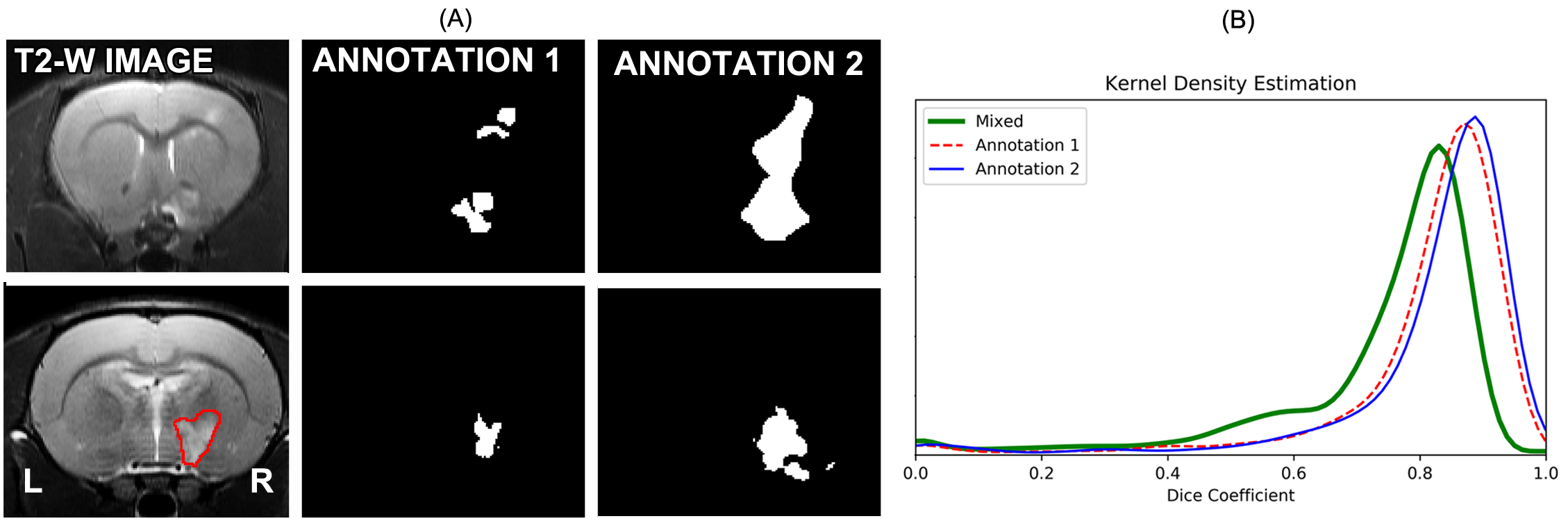}
\caption{\normalfont{a) (top row): Scan with the most disparate annotations between operators 1 and 2. a) (bottom row): A randomly selected scan of the test set (left), segmentations of the scan with RatLesNetv2 trained on Annotator 1 ground truth (middle), and Annotator 2 ground truth (right). b): Kernel density estimation of three sets of Dice coefficients. Red (dashed line) and blue (solid line) estimations were calculated between the predictions of the model optimized for the same ground truth. Green (thick solid line) estimation was computed between the predictions whose model was optimized for different ground truths. The predictions generated when the same model is optimized for different ground truths are notably different.}}
\label{fig:kde} 
\end{figure*}

As expected, optimizing separate RatLesNetv2 models with segmentations from different annotators produced more different segmentation masks than when optimizing with segmentations from the same annotator. In other words, the three sets of predictions $\hat{y}_{1,1}, \hat{y}_{1,2}, \hat{y}_{1,3}$ were similar among themselves in the same manner as $\hat{y}_{2,1}, \hat{y}_{2,2}, \hat{y}_{2,3}$ (Fig. \ref{fig:kde} b), Annotation 1 and 2), and their differences arise from the stochasticity of ConvNets optimization. In contrast, the shape of the distribution of the Dice coefficients that compare masks derived from  RatLesNetv2 models optimized with different annotations (Fig. \ref{fig:kde} b), Mixed) was notably different. Also, Annotation 1 and Mixed Dice coefficients as well as Annotation 2 and Mixed Dice coefficients were significantly different ($p$-values $<$ 0.002).

In a visual inspection, we observed that Annotation 2 was more approximate, with simpler contours, than Annotation 1. Figure \ref{fig:kde} a) (top row) shows the manual segmentations of the scan with the most disparate annotations and Fig. \ref{fig:kde} a) (bottom row) shows the predictions on a scan with the highest Dice coefficient on our baseline study when RatLesNetv2 was trained on the different annotations.

\subsection{The impact of the training set size on the performance}

\begin{figure*}[bt]
\centering
\includegraphics[width=\textwidth]{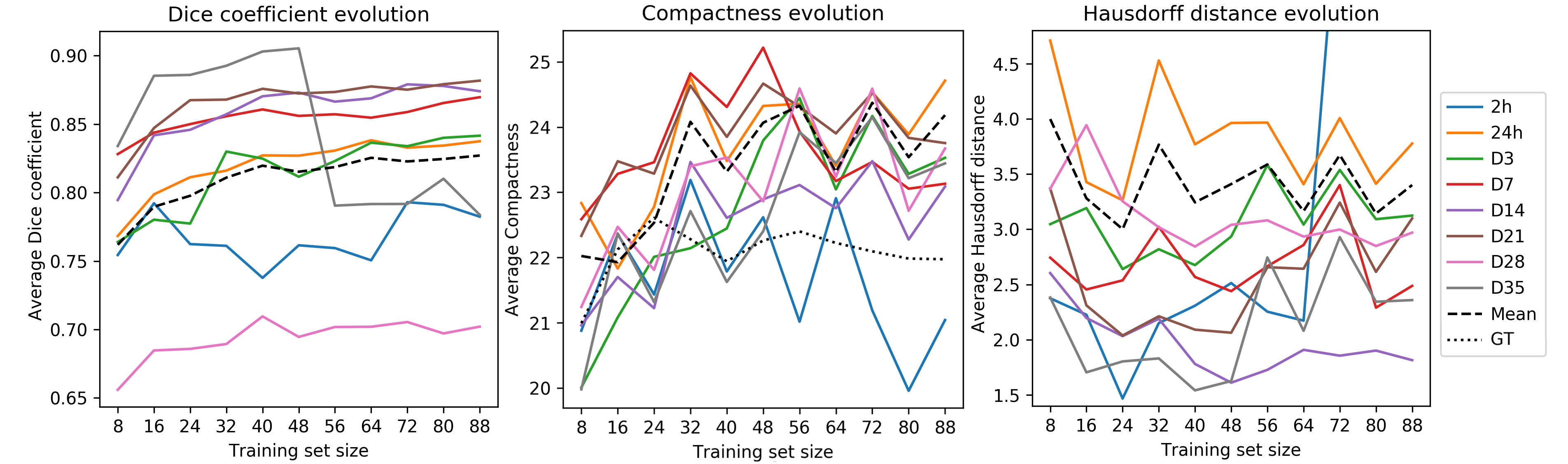}
\caption{\normalfont{RatLesNetv2 performance when optimizing for training sets of multiple sizes. Metrics (from left to right: Dice coefficient, compactness and Hausdorff distance) were processed from the masks derived with the majority voting across three runs on a fixed test set (828 images). Averages (dashed lines) were segregated by time-point. Compactness graph includes the average compactness of the ground truth (dotted line).}}
\label{fig:enlarging}
\end{figure*}

Figure \ref{fig:enlarging} illustrates the evolution of the Dice coefficients, compactness values and Hausdorff distances as the training set increases in size. Dice coefficients (Fig. \ref{fig:enlarging}, left) were remarkably different across time-points and almost every time-point reached a performance plateau with large data sets. Time-points 24h and D3---which composed the majority of the test set scans by 56.7\% and 17.8\% of the total respectively---reached their plateaus later. This effect can be a consequence of the variability within samples. On the contrary, the time-points with the lowest number of samples (2h and D35 lesions with 1 and 9 image, respectively) exhibited fluctuations.

Compactness values (Fig. \ref{fig:enlarging}, center) and Hausdorff distances (Fig. \ref{fig:enlarging}, right) oscillated considerably regardless of the time-point. Hausdorff distances were higher in the time-points with the largest number of samples (24h and D3), likely due to the existence of outliers. Compactness values, including the average (dashed line), increased analogously to the training set size, i.e., enlarging the training set yielded  less compact segmentations. Yet, these compactness values were markedly lower than the compactness values derived from segmentations produced by 3D U-Net, VoxResNet and RatLesNet (Section \ref{sec:performance}).

\section{Discussion}
We showed that RatLesNetv2 yielded similar or better Dice coefficients and Hausdorff distances, and notably more compact segmentations than other convolutional neural networks \cite{3dunet,chen2018voxresnet,valverde2019automatic}. These measurements indicate that the segmentations derived from RatLesNetv2 were more similar to the ground truth, had less large segmentation errors and were more realistic. Additionally, the smaller differences between Hausdorff distances before and after post-processing derived from RatLesNetv2 also indicate that RatLesNetv2 produced fewer segmentation errors far from the lesion surface.

RatLesNetv2 produced more compact segmentations than the other ConvNets without directly minimizing compactness (see Table \ref{table:comparisonvoxresnet}), indicating that RatLesNetv2 architecture favours segmentations with smooth borders without holes. Although optimizing compactness (and Hausdorff distance) directly might further improve the results, incorporating these terms to the loss function leads to additional hyper-parameters that require costly tuning.
Dice coefficients had large standard deviations and were lower than in existing human brain tumor segmentation studies \cite{myronenko2019robust,jiang2019two}. These results may arise due to the subjectivity of the segmentation task caused by low image contrast in certain lesions and its consequent high inter- and intra-rater disagreement. However, this is not unexpected as relatively low Dice coefficients and large standard deviations are typical in rodent \cite{mulder2017automated,valverde2019automatic} and human brain lesion segmentation studies \cite{chen2017fully,valverde2017improving,subbanna2019stroke}, even when studying inter-rater disagreement of manual annotations relying on a semi-automatic segmentation pipeline \cite{mulder2017automated}.
We also argued that Dice coefficient alone is not sufficient to measure the segmentation performance. To illustrate the importance of providing additional measurements, consider a brain with a very large and a very small lesion. If the segmentation accurately predicts the large lesion and ignores the small one, Dice coefficients will have a high value not reflecting the segmentation error, but Hausdorff distance is high capturing the segmentation error. Likewise, a lesion segmentation mask with non-smooth surface and several small holes and islands (i.e., a high compactness value) may have a high Dice coefficient despite being unrealistic.

The difference in the performance between homogeneous and heterogeneous data set divisions indicates that although few 24h lesion volumes were needed to generalize well, adding more 24h lesion volumes to the training data (homogeneous division) made RatLesNetv2 specialize on that time-point (Table \ref{table:brokendown}). On the other hand, increasing data diversity (heterogeneous division) improved performance, demonstrating that RatLesNetv2 is capable of learning from a heterogeneous data set. Thus, training on this heterogeneous division increased RatLesNetv2 capability to extrapolate to different-looking ischemic brain lesions. However, without optimizing on additional data, RatLesNetv2 performance on images with other types of lesions, such as tumor lesions, is limited by the lesions' appearance.

The ablation experiments showed that modifications of RatLesNetv2 architecture yielded similar or worse performance, justifying RatLesNetv2's architectural choices. Despite both residual connections \cite{he2016deep} and DenseNetBlocks \cite{densenet} facilitate gradient propagation \cite{drozdzal2016importance}, residual connections were preferred over DenseNetBlocks due to their notably higher performance and lower memory requirements. Additionally, a large receptive field empirically demonstrated to increase compactness and reduce large segmentation errors possibly because RatLesNetv2 considers a larger context. The choice of a large receptive field is in agreement with other state-of-the-art ConvNets that achieve large receptive fields by stacking several convolutional layers and/or utilizing dilated convolutions \cite{chen2018encoder}.

Our ground-truth disparity experiment confirmed that predictions generated when the same model is optimized for different ground truths are different. Consequently, the quality of the manually-annotated ground truth has a direct impact on the quality of the lesion masks generated automatically. As there is no unique definition of ``lesion", it may be advantageous for an algorithm to perform differently depending on the labels of the training set. On the other hand, it may also be desirable to design a robust algorithm that performs consistently regardless of some changes in the annotations.

The experiment of training RatLesNetv2 on several training sets of different sizes showed that even with few available training data RatLesNetv2 can generalize well and, despite increasing its performance when optimizing on larger training sets, such improvement is small and compactness values and Hausdorff distances fluctuate considerably.

\section{Conclusion}
We presented and made publicly available RatLesNetv2, a 3D ConvNet to segment rodent brain lesions. RatLesNetv2 has been evaluated on an exceptionally large and diverse data set of 916 rat brain MR images, validating RatLesNetv2 reliability on a wide variety of lesion stages with lesions of different appearance. Additionally, RatLesNetv2 produced segmentations that exceeded overall inter-rater agreement Dice coefficients (inter-rater: 0.73 $\pm$ 0.12, RatLesNetv2: 0.81 $\pm$ 0.16). This enhancement indicates that RatLesNetv2 produces segmentations that are remarkably more consistent with the ground truth than the similarity between different human-made annotations. This consistency is of special importance for research reproducibility, crucial in preclinical studies.

Based on our experiments and, more specifically, the accuracy greater than inter-rater agreement and than of other ConvNets, RatLesNetv2 can be used to automate lesion segmentation in preclinical MRI studies on rats.

\bibliographystyle{IEEEtran}
\bibliography{biblio}

\end{document}